\documentclass[a4paper,twoside]{article}

\usepackage{epsfig}
\usepackage{subcaption}
\usepackage{calc}
\usepackage{amssymb}
\usepackage{amstext}
\usepackage{amsmath}
\usepackage{amsthm}
\usepackage{multicol}
\usepackage{pslatex}
\usepackage{apalike}
\usepackage{algorithm2e}
\usepackage[bottom]{footmisc}
\usepackage[capitalize,nameinlink,sort&compress]{cleveref}%references for appendix
\usepackage{stfloats}%double column floats at bottom of page
\usepackage{adjustbox}%fit tables
\usepackage{multirow}
\usepackage{colortbl}%Added for color in tables
\definecolor{yellow}{rgb}{1, 1, 0.7}%2DGS
\definecolor{orange}{rgb}{1, 0.85, 0.7}%2DGS
\definecolor{tablered}{rgb}{1, 0.7, 0.7}%2DGS
\definecolor{bad}{rgb}{0.988, 0.824, 0.741}%Ours(peach puff)
\definecolor{medium}{rgb}{0.996, 1.0, 0.839}%Ours
\definecolor{good}{rgb}{0.792, 0.949, 0.761}%Ours
\definecolor{grayshade}{rgb}{0.9, 0.9, 0.9}
\newcommand{\best}{\cellcolor{good}}
\newcommand{\sbest}{\cellcolor{medium}}
\newcommand{\tbest}{\cellcolor{bad}}
\newcommand{\ignore}{\cellcolor{grayshade}}
\usepackage{SCITEPRESS}     % Please add other packages that you may need BEFORE the SCITEPRESS.sty package.

\begin{document}

% \title{Title: Subtitle}
\title{Object-Centric 2D Gaussian Splatting: Background Removal and Occlusion-Aware Pruning for Compact Object Models}

% \author{Anonymous ICPRAM submission\\
% Paper ID 120
% }

\author{\authorname{Marcel Rogge\sup{1,2} and Didier Stricker\sup{1,2}}
\affiliation{\sup{1}Augmented Vision, University of Kaiserslautern-Landau, Kaiserslautern, Germany}
\affiliation{\sup{2}Department of Augmented Vision, Deutsches Forschungszentrum fuer Kuenstliche Intelligenz, Kaiserslautern, Germany}
\email{\{marcel.rogge, didier.stricker\}@dfki.de}
}


%Abstract (70-200 words)
\abstract{Current Gaussian Splatting approaches are effective for reconstructing entire scenes but lack the option to target specific objects, making them computationally expensive and unsuitable for object-specific applications. We propose a novel approach that leverages object masks to enable targeted reconstruction, resulting in object-centric models. Additionally, we introduce an occlusion-aware pruning strategy to minimize the number of Gaussians without compromising quality. Our method reconstructs compact object models, yielding object-centric Gaussian and mesh representations that are up to 96\% smaller and up to 71\% faster to train compared to the baseline while retaining competitive quality. These representations are immediately usable for downstream applications such as appearance editing and physics simulation without additional processing.}

% \onecolumn \maketitle \normalsize \setcounter{footnote}{0} \vfill
\twocolumn[{%
\renewcommand\twocolumn[1][]{#1}%
\maketitle
\begin{center}
    \centering
    \vspace{-1cm}
    \captionsetup{type=figure}
    \includegraphics[width=\textwidth]{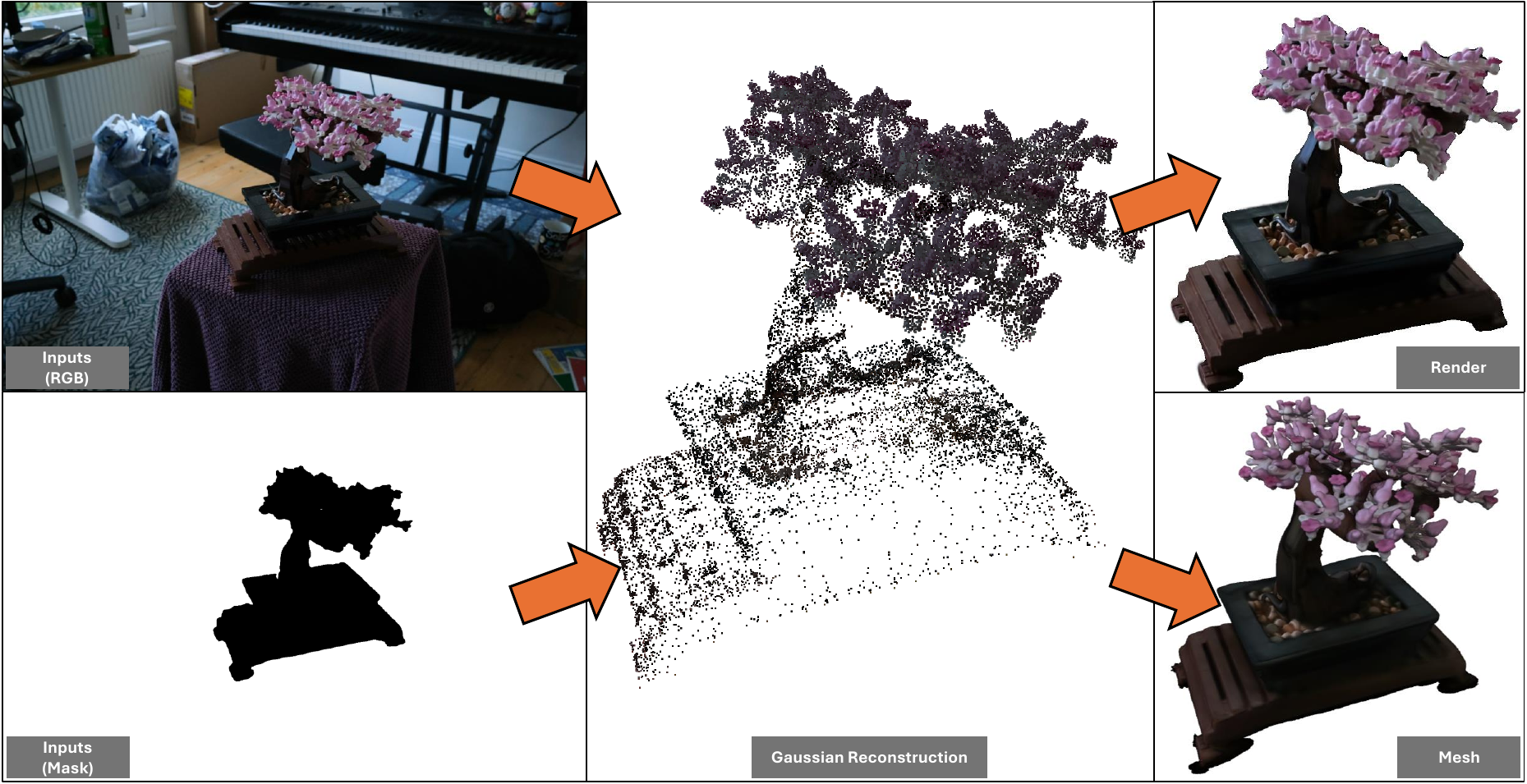}
    \captionof{figure}{Our method, optimizes 2D Gaussians to accurately model specific object surfaces. They can be rendered directly or exported as a mesh. For ease of viewing, the mask is inverted and the rendered image's background is edited to be white.}
    \label{fig:teaser}
    \vspace{0.5cm}
\end{center}%
}]

\normalsize \setcounter{footnote}{0} \vfill

%Introduction
\section{\uppercase{Introduction}}

Multi-view 3D reconstruction has seen a surge of attention in recent years. 
The introduction of Neural Radiance Fields (NeRF)~\cite{mildenhall2020nerf} has made the high-quality reconstruction of complex scenes possible. 
However, the implicit nature of NeRFs makes it difficult to utilize the underlying scene representation. 
This motivates the extraction of explicit representations~\cite{yariv2023bakedsdf}. 
More recently, the explicit method 3D Gaussian Splatting (3DGS)~\cite{kerbl3Dgaussians} shows high-quality rendering results and fast rendering speeds. 
Its explicit nature makes it easier to utilize in downstream applications such as visualization and editing. 
However, it is a new type of representation, which requires custom rendering software to display it correctly. 
Recent works convert Gaussian representations into meshes~\cite{Huang2DGS2024,guedon2023sugar}, which enables support for traditional applications. 
However, these methods are inefficient when only specific objects need to be reconstructed, as they reconstruct anything that is visible in the input images. 
Instead, our proposed method uses a novel background loss to remove background Gaussians as defined by a segmentation mask. 
This improves efficiency by reducing the training time and model size. 
We further reduce the model size without any loss in quality by utilizing an occlusion-aware pruning strategy which removes Gaussians that do not contribute to the rendering. 
Our method produces high-quality Gaussian and mesh representations (Figure~\ref{fig:teaser}), which can immediately be used for downstream applications (\cref{appendix:downstream_applications}).

The contributions are the following:
\begin{itemize}
    \item A background loss that enables object-centric reconstruction, which speeds up training and significantly reduces the model size.
    \item A general pruning strategy for Gaussian-based methods to remove occluded Gaussians that do not contribute to the overall scene representation.
    \item We achieve competitive reconstruction quality for object-centric Gaussian and mesh models that can immediately be used in downstream applications.
\end{itemize}

%Related Work
\section{\uppercase{Related Work}}

\subsection{Novel View Synthesis}

The area of novel view synthesis has seen significant advancements since the release of NeRF~\cite{mildenhall2020nerf}. 
NeRF shows that we can encode an implicit representation of a scene into a multi-layer perceptron (MLP). 
The MLP is trained using traditional volume rendering techniques to render images from known poses. 
By optimizing the resulting rendered views to match the ground truth images, the MLP learns a 3D consistent geometry due to the used volume rendering. 
Since then, many methods have further improved the quality of NeRF by changing the sampling technique~\cite{barron2021mipnerf} and enabling the use of unbounded scenes~\cite{barron2022mipnerf360}. 
However, while these NeRFs are able to produce high-quality novel views, they are prohibitively expensive making real-time rendering impossible. 
Other works focus therefore on making it possible to render NeRFs in real time by baking the trained models into new representations~\cite{hedman2021snerg,Reiser2023merf}.

The introduction of 3DGS~\cite{kerbl3Dgaussians} offers a new approach to tackle novel view synthesis after NeRF. 
3DGS does not rely on an MLP and instead optimizes a discrete set of three-dimensional Gaussians. 
Optimization is possible through a differentiable rendering approach that utilizes tile-based rasterization. 
It manages to achieve a quality that is close to the best NeRF methods while being significantly faster due to the efficient rendering approach. 
This includes training the models faster and also real-time rendering of the trained models. 
However, 3DGS introduces a novel scene representation that cannot be visualized or manipulated using traditional software. 
Therefore, recent works try to expand the Gaussian representation with additional features to support e.g. relighting~\cite{R3DG2023relightable} and physics simulations~\cite{xie2023physgaussian}. 
Our method also uses a Gaussian representation but offers support for traditional software by enabling the extraction of a mesh.

\subsection{Mesh Reconstruction}

Traditional scene representations like meshes have been around for a long time and benefit from strong support in computer graphics applications. 
Therefore, work has been published on converting the previously mentioned representations into meshes. 
One method modifies an underlying NeRF model to better learn surfaces which are then baked into a mesh~\cite{yariv2023bakedsdf}. 
The authors showcase various uses of their extracted meshes for downstream tasks like physics simulation and appearance editing. 
Other works, such as Neus~\cite{wang2021neus} and VolSDF~\cite{yariv2021volsdf}, directly optimize an implicit signed distance function (SDF), which makes it possible to obtain high-quality meshes. 
Finally, there are also works that convert the explicit Gaussian representations into meshes. 
SuGaR~\cite{guedon2023sugar} encourages 3D Gaussians to align themselves with surfaces in the scene, which can then be used to extract a mesh.

Another approach is 2D Gaussian Splatting (2DGS)~\cite{Huang2DGS2024}, where the three-dimensional Gaussians are replaced by two-dimensional Gaussian discs. 
These Gaussian discs are suited to model surfaces and are additionally encouraged through regularization to gather closely together to model surfaces of the scene. 
This ensures high-quality depth renders of the scene, which are then used to extract a mesh. 
However, it is difficult to create a mesh of a specific object in the scene from the Gaussian representation that encompasses the whole scene. 
This can only reliably be achieved with the use of segmentation masks during the mesh extraction. 
Using an available mask only during the mesh creation is, however, wasteful because a lot of computing power is used to reconstruct the entire scene. 
Our method solves this by involving available masks directly during the scene optimization. 
This reduces compute resources and additionally makes mesh generation easier by removing the need to make assumptions about the scene bounds during the extraction.

\subsection{Object Reconstruction}

The concurrent work GaussianObject~\cite{yang2024gaussianobject} also considers object-centric reconstruction. 
However, their approach fundamentally differs from ours. 
First, they consider very-sparse-view reconstruction while we consider a standard multi-view reconstruction scenario. 
Although their approach handles a more difficult scenario, they solve it through special preprocessing and a multi-step training approach. 
The authors do not publish individual training times but indicate an approximate time that is more than three times slower than ours while using images at half the resolution. 
Second, they utilize 3D Gaussians while we use 2D Gaussians. 
The 2D Gaussian representation is better suited to extract meshes, which improves the usability of our method for downstream applications~\cite{Huang2DGS2024}.

%Method
\section{\uppercase{Preliminaries}}

\subsection{Motivation}\label{sec:preliminaries_motivation}

%Our method aims to address the shortcomings of the various novel view synthesis approaches. 
NeRF models tend to achieve very high accuracy but are slow to train and render~\cite{kerbl3Dgaussians}. 
Follow-up works, such as SNeRG~\cite{hedman2021snerg}, convert NeRF models into representations that are faster to render but the initial training step remains slow. 
Gaussian models are fast to train and render images in real time while still producing excellent quality \cite{kerbl3Dgaussians}. 
However, the Gaussian representation requires custom rendering software that is not yet widely available. 
Additionally, modifying the Gaussian representation or rendering pipeline, such as is the case with 3DGS~\cite{kerbl3Dgaussians} and 2DGS~\cite{Huang2DGS2024}, requires corresponding modifications in any downstream application that supports Gaussians. 
The option to extract meshes ensures support for downstream tasks such as appearance editing and physics simulations using existing applications~\cite{yariv2023bakedsdf}. 
We will consider 2DGS as a foundation because they show fast train and render times with high-quality rendering, as well as the option to extract meshes from the underlying representation.

\subsection{2D Gaussian Splatting}\label{sec:preliminaries_2dgs}

%2DGS presented some of the best quality in terms of rendered views and meshes, as well as some of the lowest training times and memory requirements. 
%For this reason, we are using 2DGS as a base for our proposed method though our contributions can easily be included into any gaussian-based method. 
We motivate our choice of 2DGS as the base for our proposed method in \cref{sec:preliminaries_motivation}, although our contributions can be included in most Gaussian-based methods. 
We validate this on the original 3DGS in \cref{appendix:3dgs}. 
In the following, we will detail the general steps of 2DGS, which will make the impact of our contributions clearer. 
Figure~\ref{fig:pipeline_overview} also provides an overview including our proposed changes, which will be discussed in \cref{sec:method}. 
We refer to the 2DGS paper as well as the previous 3DGS paper for more specifics about the underlying principles.

\subsubsection{Optimizing the Gaussian Representation}

Given a set of $n$ unstructured images $I_i\in \mathbb{R}^{h\times w\times 3}$, where $i\in \{1, ..., n\}$ and $h,w$ are the height and width of the image, the goal is to create a geometrically accurate 3D reconstruction that enables fast rendering of novel views. 
In a preprocessing step, Structure from Motion (SfM) is performed to obtain the $SE(3)$ camera poses $P_i\in \mathbb{R}^{4\times 4}$ corresponding to each input image $I_i$. 
Additionally, SfM outputs a sparse point cloud of the scene as a by-product of the pose estimation. 
The sparse point cloud is used as a meaningful initialization by creating a 2D Gaussian for each point. 
From here, the optimization loop starts: each iteration, one of the input views $I_i$ is selected. 
Given the input's pose $P_i$, a view $R_i\in \mathbb{R}^{h\times w\times 3}$ of the 2D Gaussian representation is rendered. 
Then, a photometric loss is computed between the rendered view $R_i$ and the ground truth $I_i$. 
Additionally, two regularization terms are computed for the depths and normals of the Gaussians used to render $R_i$. 
Our proposed method adds one additional loss term on the opacity of the Gaussians as detailed in \cref{sec:method:background_loss}. 
Through backpropagation, the position, shape, and appearance of all involved Gaussians are updated to better fit the color of $I_i$. 
Lastly, there is an adaptive densification control which duplicates and prunes Gaussians in set intervals based on their accumulated gradients. 
We propose to expand the adaptive densification control with the pruning of occluded Gaussians which we detail in \cref{sec:method:pruning_strategy}. 
After optimizing for a set number of iterations, all 2D Gaussians are exported as the 3D representation of the scene. 
While 2DGS optimizes an entire scene, our proposed method learns a target object and exports its isolated 3D representation (\cref{sec:method}).

\subsubsection{Extracting Meshes}

Optionally, it is possible to extract a mesh from the 2D Gaussian representation. 
First, the scene is rendered from each of the training poses $P_i$, returning the colors $R_i$ and depths $D_i\in \mathbb{R}^{h\times w\times 1}$. 
From here, there are two options: 'bounded' and 'unbounded' mesh extraction. 
The bounded setting truncates the 3D representation based on a depth value $d_{trunc}$, which is either empirically set or automatically estimated based on the scene bounds. 
Using Open3d~\cite{Zhou2018open3d}, a Truncated Signed Distance Function (TSDF) volume is created from all $R_i$ and $D_i$, where $D_i$ is truncated based on $d_{trunc}$. 
Finally, Open3D can convert the TSDF volume into a triangle mesh. 
The unbounded setting instead considers the entire 3D representation without any depth truncation by contracting everything into a sphere. 
The authors then utilize a customized TSDF computation and marching cube algorithm to obtain a mesh.

2DGS optionally considers segmentation masks $M_i\in \mathbb{R}^{h\times w\times 1}$ of an object $O$, with $M_i$ equals 1 for pixels in $I_i$ that show $O$ and $M_i$ equals 0 otherwise. 
These segmentation masks are used during the mesh extraction only if using the bounded setting. 
If $M_i$ is available, before integration into the TSDF volume, $D_i$ will additionally be truncated for pixels that do not show $O$, i.e. $M_i$ equals 0. 
In \cref{sec:implementation}, we will make a distinction for 2DGS based on whether $M_i$ was used to extract a mesh or not.

\begin{figure*}[t]
  \centering
   {\epsfig{file = 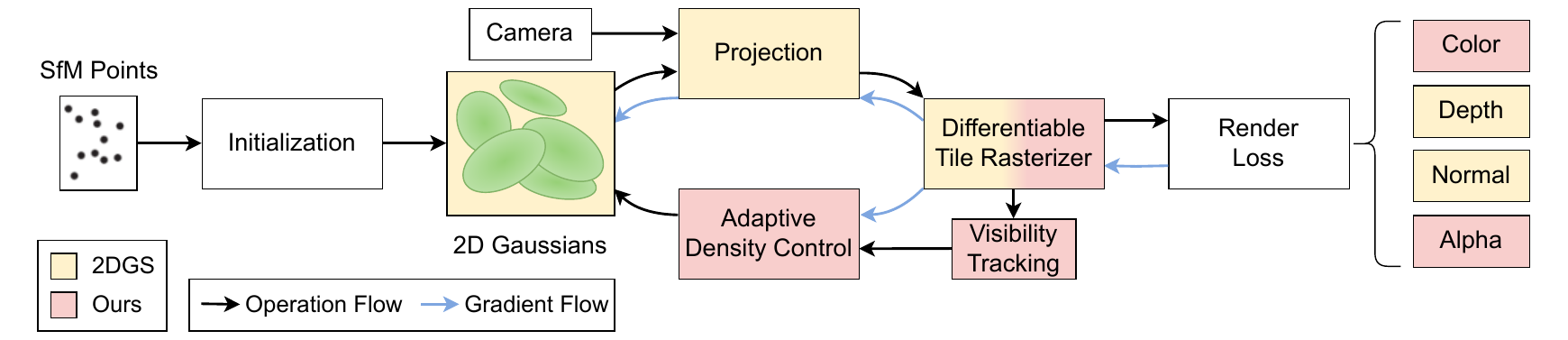, width = \linewidth}}
  \caption{Overview of our method adapted from the 3DGS paper~\cite{kerbl3Dgaussians}. Changes to the original pipeline are highlighted for 2DGS and Ours. Overall, 2D Gaussians are initialized using a sparse SfM point cloud. During optimization, the density of the Gaussians is adaptively controlled. The rasterization-based renderer enables very fast training and inference.}
  \label{fig:pipeline_overview}
\end{figure*}

\section{\uppercase{Compact Object Reconstruction}}\label{sec:method}

We use 2DGS as a base and expand it with our background loss (\cref{sec:method:background_loss}) and our pruning strategy for occluded Gaussians (\cref{sec:method:pruning_strategy}). 
Our background loss utilizes segmentation masks $M_i$ to make the reconstruction object-centric, which reduces training time and the size of the model. 
Our pruning strategy removes unnecessary Gaussians, which reduces the model size further without any loss in quality. 
Our background loss requires masks $M_i$, which are not always available. 
\cref{sec:method:mask_generation} details how we generate masks if they are not available. 
An overview of the method is shown in Figure~\ref{fig:pipeline_overview}, where contributions are highlighted.

\subsection{Mask Generation}\label{sec:method:mask_generation}

\begin{figure}[hb]
  \centering
   {\epsfig{file = 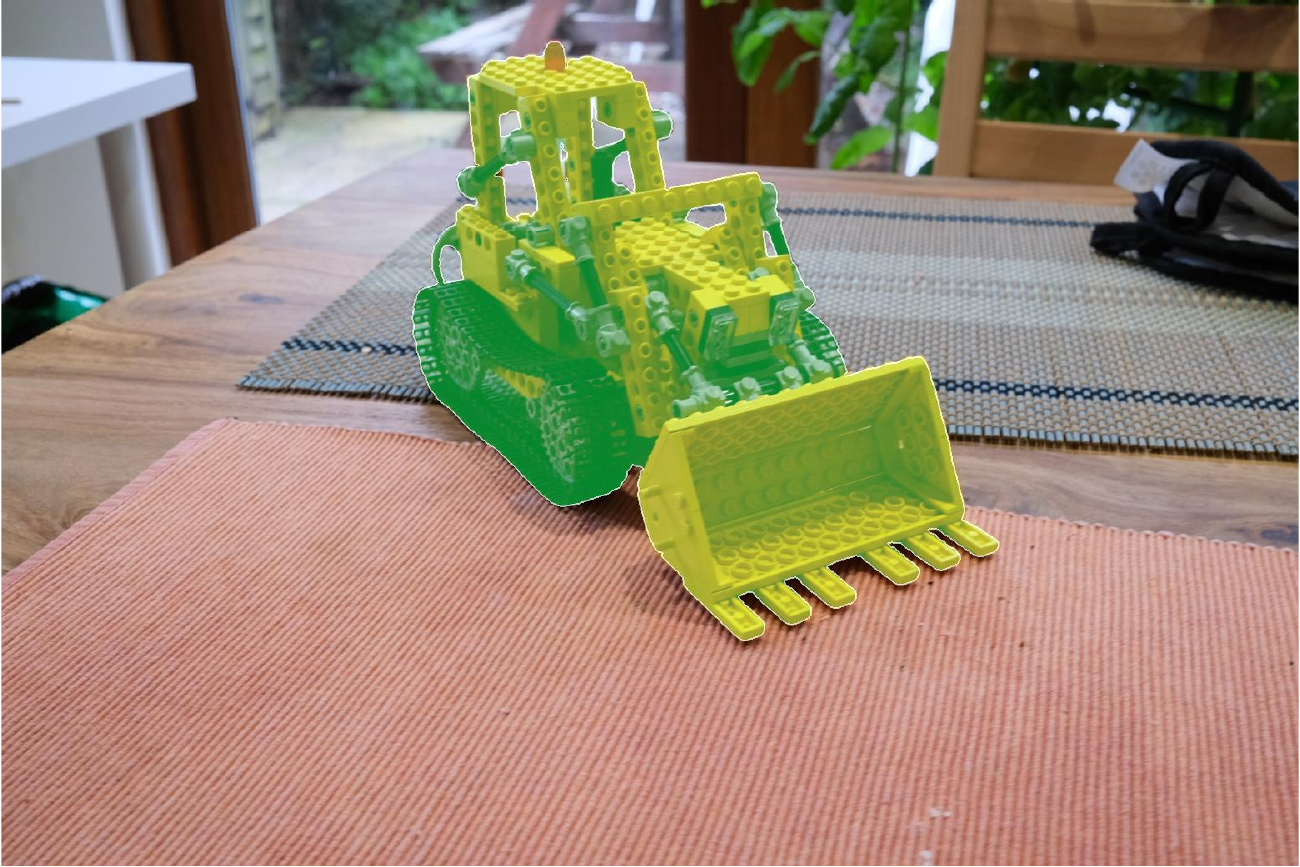, width = \linewidth}}
  \caption{Mask from SAM~2 overlaid on the input image.}
  \label{fig:mask_overlay}
\end{figure}

Recent advances in image segmentation have made the creation of object masks relatively easy. 
Using Segment Anything 2 (SAM~2)~\cite{ravi2024sam2}, it is possible to semi-automatically generate masks of specific objects across an image sequence. 
Usually, it is possible to generate accurate masks by interactively setting only a few markers on an image. 
Objects segmented in one image can then be propagated through an entire image sequence. 
The accuracy is dependent on the complexity of the image sequence and the characteristics of the target object. 
If necessary, the segmentation results can be refined by adding markers on additional images and propagating the new result. 
We utilize SAM~2 to generate object masks for the scenes of the Mip-NeRF360 dataset~\cite{barron2022mipnerf360}, which does not provide any masks. 
An example mask is shown in Figure~\ref{fig:mask_overlay}.

\subsection{Background Loss}\label{sec:method:background_loss}

2DGS is able to produce high-quality meshes of objects if segmentation masks are available. 
However, the learned Gaussian representation consists of the entire scene and the object is only isolated during the mesh generation. 
This wastes computation resources by reconstructing parts of the scene which are discarded afterwards. 
Additionally, the underlying Gaussians are not representative of the target object. 
This makes them less suitable for applications that are able to work directly with Gaussian representations because the object must first be isolated.

Our goal is to mask objects during the optimization loop of 2DGS in order to reconstruct only the necessary parts of the scene. 
Formally, given images $I_i$ and corresponding segmentation masks $M_i$ of an object $O$, let $M_i$ be 1 for pixels in $I_i$ where $O$ is present and 0 otherwise. 
When rendering the view $R_i$, the rasterization of Gaussians should result in no Gaussians where $M_i$ is 0. 
We formulate our background loss as follows:
\begin{equation}\label{eq:loss_background}
    \mathcal{L}_{b} = \frac{1}{h\cdot w} \sum[A_i \cdot (1 - M_i)],
\end{equation}
where $A_i\in \mathbb{R}^{h\times w\times 1}$ represents the accumulated alphas from Gaussians during rendering of $R_i$. 
This is highlighted in our pipeline as render loss on 'Alpha' (Figure~\ref{fig:pipeline_overview}). 
Our background loss effectively penalizes Gaussian opacity where $O$ is not present, pushing them to become transparent. 
Transparent Gaussians are automatically pruned via a threshold as part of the 3DGS density control.

The proposed background loss is, however, at odds with the 3DGS photometric loss which pushes the Gaussians to match the colors of $R_i$ and $I_i$ even where $M_i$ is 0. 
To avoid training of unwanted parts of the image, we multiply both $I_i$ and $R_i$ with $M_i$ to compute the masked photometric loss $\mathcal{L}_c^M$.
\begin{align}\label{eq:loss_photometric}
    \mathcal{L}_{c}^{M}(I_i, R_i, M_i) &=  \mathcal{L}_{c}(I_i \cdot M_i, R_i \cdot M_i),
\end{align}
where $\mathcal{L}_c$ is the photometric loss from Equation~7 in the 3DGS paper~\cite{kerbl3Dgaussians}. 
We highlight this in our pipeline as render loss on 'Color' (Figure~\ref{fig:pipeline_overview}). 
The total loss is as follows:
\begin{equation}\label{eq:loss_total}
    \mathcal{L} = \mathcal{L}_{c}^{M} + \alpha\mathcal{L}_{d} + \beta\mathcal{L}_{n} + \gamma\mathcal{L}_{b},
\end{equation}
where $\mathcal{L}_{d}$ is the 2DGS depth distortion loss and $\mathcal{L}_{n}$ is the 2DGS normal consistency loss. 
It is not necessary to perform any masking on the two terms from 2DGS since they do not influence the opacity of Gaussians.

\subsection{Pruning Strategy}\label{sec:method:pruning_strategy}

We introduce a novel pruning approach that tracks which Gaussians have been used during rendering, i.e., Gaussians that are involved in the alpha blending of any rendered pixel. 
We will refer to this concept from here on as 'visible' Gaussians or Gaussians that are not 'occluded'. 
The concept is similar to the 'visibility filter' used as part of the 3DGS density control, which tracks Gaussians that are inside the viewing frustum for a given view. 
However, there is a significant difference between a Gaussian that is actually visible and a Gaussian being in the field of view. 
Formally, given $m$ Gaussians, let $G_j$ be a Gaussian that is inside the viewing frustum and $G_k$ a Gaussian that is outside the viewing frustum, where $j,k\in \{1, ..., m\}$ and $j\neq k$. 
During preprocessing, $G_j$ is projected to the image and rasterized, identifying which pixels it can contribute to. 
At the same time, $G_k$ is disqualified from rendering and not assigned to any pixel. 
During the rendering of a pixel $p$, all assigned Gaussians are rendered in front-to-back order, accumulating their colors and opacities to $p$. 
Rendering is terminated once either $p$ reaches a threshold opacity close to 1 or all assigned Gaussians have been exhausted. 
Let us assume that all pixels that $G_j$ is assigned to, reach the threshold opacity before reaching $G_j$ during rendering. 
The 3DGS visibility filter will assign $G_j$ with true and $G_k$ with false. 
We propose to assign both $G_j$ and $G_k$ with false.
%Specifically, the visibility filter will be true for gaussians that are fully occluded by other gaussians, such as an object occluded by a second object or even the backside of a visible object. 
%Whereas, the seen status of a gaussian will only be true if there are no gaussians occluding it or the occluding gaussians do not have sufficient opacity to reach the termination threshold of the renderer.

The motivation to track the visibility is clear: Gaussians that are occluded cannot receive gradients and no longer participate in the optimization process. 
At the same time, these Gaussians are meaningless to the overall scene representation since they exist in a space that is not observed. 
% While a single Gaussian is insignificant in terms of memory, the number of times that this occurs with scenes comprising millions of Gaussians is significant. 
A few Gaussians do not occupy a lot of memory and require only a few computations during preprocessing. 
However, with scenes comprising millions of Gaussians, a substantial amount of resources is wasted. 
We validate the magnitude of occluded Gaussians in \cref{sec:exp_m360}.

%Experiments
\section{\uppercase{Experiments}}\label{sec:experiments}

We present evaluations of our proposed method and its 2DGS baseline. 
For completeness, we also present some results from previous state-of-the-art methods that were compared with in the 2DGS paper~\cite{Huang2DGS2024}. 
Lastly, we will evaluate the efficacy of each component of our method.

\subsection{Implementation}\label{sec:implementation}

We extend the custom CUDA kernels of 2DGS, which are built upon the original 3DGS framework. 
Inside the renderer, we save which Gaussians are used during the color accumulation step. 
Regarding the adaptive density control, we use the same default values as 2DGS, which correspond to the 3DGS defaults at that time. 
The pruning of occluded Gaussians is performed at an interval of 100 and 600 iterations for the DTU~\cite{jensen2014dtu} and Mip-NeRF360 dataset~\cite{barron2022mipnerf360}, respectively. 
The loss term coefficients as introduced by 3DGS and 2DGS are unchanged. 
Although the inputs to the photometric loss are masked as described in \cref{sec:method:background_loss}. 
Our background loss utilizes alpha maps returned from the renderer. 
We set $\gamma$ to 0.5 for the background loss in all our experiments, unless stated otherwise. 
All experiments on our method and 2DGS are run with 30k iterations on an RTX3090 GPU.

\textbf{Mesh Extraction}. 
We use the 2DGS approach detailed in \cref{sec:preliminaries_2dgs} to extract meshes from the learned Gaussian representation. 
Notably, we observe that 2DGS uses masks during the extraction process of the mesh. 
However, we believe this would be unfair when comparing with other methods that generate meshes without using masks. 
For this reason, we will consider our 2DGS results in two settings: 1) With masks, and 2) Without masks. 
Where the first case is equivalent to the standard 2DGS pipeline and in the second case we skip the use of masks during the mesh extraction. 
The use of masks to cull the final mesh for the computation of the error metric is valid in both scenarios. 
This is to isolate the error related to the target object only. 
We will compare the results of our proposed method and previous state-of-the-art methods based on the use of masks accordingly. 
For evaluations on the DTU dataset, we used the same values for voxel size and truncation thresholds as 2DGS to enable a fair comparison. 
The only deviation is given for our approaches with masks: we do not cull the masks before computation of the error metric because our mesh already consists of only the target object. 
Meshes generated from the Mip-NeRF360 dataset are only for qualitative comparison. 
Therefore, we change the mesh extraction for our method (with masks) to the unbounded mode with the voxel size fixed to $0.004$. 
Since our reconstructed scene consists only of the object, we can safely turn the entire scene into a mesh to obtain the object's mesh. 
Notably, unlike the bounded setting, there is no need to estimate any boundaries, making the unbounded setting the most generalizable. 
Theoretically, it should work for any object reconstruction using our proposed method without the need to tweak any parameters.

\subsection{Datasets}

We evaluate our method on the DTU~\cite{jensen2014dtu} and Mip-NeRF360~\cite{barron2022mipnerf360} datasets. 
DTU consists of a subset of 15 scenes from a larger dataset which are widely used for evaluation by 3D reconstruction methods. 
Each of the 15 scenes consists of either 49 or 64 images. 
Point clouds obtained from a structured light scanner serve as ground truth for 3D reconstructions. 
We use the dataset provided by 2DGS, which already combines the RGB images with object masks and has the necessary sparse point cloud obtained from Colmap~\cite{schoenberger2016sfm,schoenberger2016mvs} available. 
Additionally, we download the ground truth point clouds from the DTU authors for computation of the error metric. 
All experiments were performed at the same 800~$\times$~600 resolution as chosen by the 2DGS authors.

Mip-NeRF360 consists of 9 scenes; 5 outdoor and 4 indoor. 
The outdoor and indoor scenes were captured with two cameras at resolutions of 4946~$\times$~3286 and 3118~$\times$~2078, respectively. 
Each scene consists of between 125 and 311 images and a sparse point cloud obtained from Colmap. 
All experiments on outdoor scenes were performed at a resolution of 1237~$\times$~822, which roughly represents a downsampling by factor 4. %full-res 4946x3286
Experiments on the indoor scenes were performed at a resolution of 1559~$\times$~1039, which represents a downsampling by factor 2. %full-res 3118x2078
These image resolutions are the same as specified in the metric evaluation scripts available from 3DGS and 2DGS.

\subsection{Evaluations}

\begin{table*}
    \caption{Quantitative comparison on the DTU dataset~\cite{jensen2014dtu} for methods with using masks. Results for methods marked with $\dagger$ and ${\ddagger}$ are taken from the 2DGS~\cite{Huang2DGS2024} and Neus~\cite{wang2021neus} papers, respectively. All others are the results from our experiments. Time is indicated as hours or minutes. All others are chamfer distance. Green indicates the best, yellow indicates the second-best, and orange indicates the third-best result.}
    \label{tab:DTU_w/_mask}
    \adjustbox{max width=\textwidth}{
        \centering
        \begin{tabular}{|ll|c|c|c|c|c|c|c|c|c|c|c|c|c|c|c|c|c|}
            \hline
            & \textbf{w/ mask} & 24 & 37 & 40 & 55 & 63 & 65 & 69 & 83 & 97 & 105 & 106 & 110 & 114 & 118 & 122 & Mean & Time \\
            \hline
            \multirow{2}{*}{\rotatebox{90}{impl.}} & NeRF$^{\ddagger}$ & 1.83 & 2.39 & 1.79 & 0.66 & 1.79 & 1.44 & 1.50 & \best 1.20 & 1.96 & 1.27 & 1.44 & 2.61 & 1.04 & 1.13 & 0.99 & 1.54 & N/A \\%From Neus but without remark on time
            & Neus$^{\ddagger}$ & \tbest 0.83 & \sbest 0.98 & \tbest 0.56 & \best 0.37 & \tbest 1.13 & \best 0.59 & \best 0.60 & 1.45 & \best 0.95 & \tbest 0.78 & \best 0.52 & \tbest 1.43 & \best 0.36 & \best 0.45 & \best 0.45 & \sbest 0.77 & $\sim$14h \\
            \hline
            \multirow{5}{*}{\rotatebox{90}{explicit}} & 3DGS$^{\dagger}$ & 2.14 & 1.53 & 2.08 & 1.68 & 3.49 & 2.21 & 1.43 & 2.07 & 2.22 & 1.75 & 1.79 & 2.55 & 1.53 & 1.52 & 1.50 & 1.96 & \tbest 11.2m \\
            & SuGaR$^{\dagger}$ & 1.47 & \tbest 1.33 & 1.13 & 0.61 & 2.25 & 1.71 & 1.15 & 1.63 & 1.62 & 1.07 & 0.79 & 2.45 & 0.98 & 0.88 & 0.79 & \tbest 1.33 & $\sim$1h \\
            & 2DGS$^{\dagger}$ & \ignore 0.48 & \ignore 0.91 & \ignore 0.39 & \ignore 0.39 & \ignore 1.01 & \ignore 0.83 & \ignore 0.81 & \ignore 1.36 & \ignore 1.27 & \ignore 0.76 & \ignore 0.70 & \ignore 1.40 & \ignore 0.40 & \ignore 0.76 & \ignore 0.52 & \ignore 0.80 & \ignore 10.9m \\
            & 2DGS & \best 0.45 & \best 0.82 & \sbest 0.31 & \sbest 0.38 & \sbest 0.95 & \sbest 0.83 & \sbest 0.80 & \tbest 1.30 & \tbest 1.16 & \best 0.68 & \sbest 0.66 & \sbest 1.36 & \sbest 0.39 & \sbest 0.66 & \sbest 0.48 & \best 0.75 & \sbest 10.94m \\
            & Ours & \sbest 0.47 & \best 0.82 & \best 0.30 & \tbest 0.43 & \best 0.93 & \tbest 0.96 & \tbest 0.86 & \sbest 1.26 & \sbest 1.03 & \sbest 0.72 & \tbest 0.74 & \best 1.25 & \tbest 0.47 & \tbest 0.75 & \tbest 0.55 & \sbest 0.77 & \best 6.46m \\
            \hline
        \end{tabular}
    }
\end{table*}
\begin{table*}
    \caption{Quantitative comparison on the DTU dataset~\cite{jensen2014dtu} for methods without using masks. Results for methods marked with $\dagger$ and ${\ddagger}$ are taken from the VolSDF~\cite{yariv2021volsdf} and Neus~\cite{wang2021neus} papers, respectively. All others are the results from our experiments. Time is indicated as hours or minutes. All others are chamfer distance. Green indicates the best, yellow indicates the second-best, and orange indicates the third-best result.}
    \label{tab:DTU_w/o_mask}
    \adjustbox{max width=\textwidth}{
        \centering
        \begin{tabular}{|ll|c|c|c|c|c|c|c|c|c|c|c|c|c|c|c|c|c|}
            \hline
            & \textbf{w/o mask} & 24 & 37 & 40 & 55 & 63 & 65 & 69 & 83 & 97 & 105 & 106 & 110 & 114 & 118 & 122 & Mean & Time \\
            \hline
            \multirow{3}{*}{\rotatebox{90}{implicit}} & NeRF$^{\ddagger}$ & 1.90 & 1.60 & 1.85 & 0.58 & 2.28 & 1.27 & 1.47 & 1.67 & 2.05 & 1.07 & 0.88 & 2.53 & 1.06 & 1.15 & 0.96 & 1.49 & N/A \\%Listed in Neus paper, comes from UNISURF, no remark on time
            & Neus$^{\ddagger}$ & \tbest 1.00 & 1.37 & \tbest 0.93 & \sbest 0.43 & \tbest 1.10 & \best 0.65 & \best 0.57 & 1.48 & \best 1.09 & 0.83 & \best 0.52 & \sbest 1.20 & \best 0.35 & \best 0.49 & \sbest 0.54 & \tbest 0.84 & $\sim$16h \\
            & VolSDF$^{\dagger}$ & 1.14 & \tbest 1.26 & \sbest 0.81 & \tbest 0.49 & 1.25 & \sbest 0.70 & \sbest 0.72 & \best 1.29 & \sbest 1.18 & \best 0.70 & \sbest 0.66 & \best 1.08 & 0.42 & \sbest 0.61 & \tbest 0.55 & 0.86 & \tbest $\sim$12h \\
            \hline
            \multirow{2}{*}{\rotatebox{90}{expl.}} & 2DGS & \best 0.47 & \sbest 0.89 & \best 0.37 & \best 0.39 & \best 0.95 & 0.85 & \tbest 0.82 & \sbest 1.40 & \sbest 1.18 & \tbest 0.78 & \tbest 0.67 & 1.37 & \sbest 0.39 & 0.67 & \best 0.52 & \best 0.78 & \sbest 10.94m \\
            & Ours & \sbest 0.49 & \best 0.88 & \best 0.37 & \best 0.39 & \sbest 1.01 & \tbest 0.83 & \tbest 0.82 & \tbest 1.41 & \tbest 1.27 & \sbest 0.76 & 0.71 & \tbest 1.28 & \tbest 0.41 & \tbest 0.66 & \sbest 0.54 & \sbest 0.79 & \best 10.76m \\
            \hline
        \end{tabular}
    }
\end{table*}

\subsubsection{DTU}

We evaluate the quality of mesh reconstruction on the DTU dataset using the chamfer distance. 
Here we make a distinction between methods with masks and methods without masks. 
In regards to 2DGS, we detail this distinction in \cref{sec:implementation}. 
Results from methods other than 2DGS and ours have been indicated from which paper they have been taken. 
Our proposed method, in the case of without masks, is equivalent to only using the proposed pruning strategy~\cref{sec:method:pruning_strategy}.

The evaluation of quality and training time for the case of with masks can be found in Table~\ref{tab:DTU_w/_mask}. 
We observe that our proposed method produces an equivalent quality to the best implicit method Neus~\cite{wang2021neus} while being 100$\times$ faster. 
Compared to the explicit methods, our method has a minor drop in quality compared to 2DGS but is almost twice as fast. 
We would also like to highlight that our method does not require the mesh to be culled before the computation of the loss and indicates the quality of the entire mesh. 
The discrepancy between the results reported in the 2DGS paper and our own experiments is due to improvements to the code that the authors have released since the publication of their paper.

The evaluation of quality and training time for the case without using masks can be found in Table~\ref{tab:DTU_w/o_mask}. 
We observe that there is almost no difference in quality between our proposed method and 2DGS. 
This is because our pruning strategy removes only unnecessary Gaussians. 
At the same time, the reduced number of Gaussians positively impacts the speed of our method, making it a little faster. 
We would like to note that the gain in speed is not that significant due to the occluded Gaussians not taking part in the rendering itself. 
For this reason, we will also evaluate the number of Gaussians separately. 
Compared to the implicit methods, both our proposed method and 2DGS produce higher-quality meshes while being significantly faster.

\begin{table}
    \centering
    \scriptsize
    \caption{Performance comparison on the DTU dataset. Methods marked with $\star$ are without using masks.}
    \label{tab:DTU_performance}
    \begin{tabular}{|l|c|c|c|c|}
        \hline
        & CD$\downarrow$ & Time$\downarrow$ & Gaussians$\downarrow$ & Storage$\downarrow$ \\
        \hline
        2DGS$^\star$ & \tbest 0.782 & \tbest 10.94 min & \tbest 198,820 & \tbest 46.21 MB \\
        %\hline
        Ours$^\star$ & 0.787 & \sbest 10.76 min & \sbest 178,332 & \sbest 41.45 MB \\
        \hline
        2DGS & \best 0.748 & \tbest 10.94 min & \tbest 198,820 & \tbest 46.21 MB \\
        %\hline
        Ours & \sbest 0.769 & \best 6.46 min & \best 108,568 & \best 25.23 MB \\
        \hline
    \end{tabular}
\end{table}

Table~\ref{tab:DTU_performance} gives an overview of the performance of our proposed method and 2DGS in two settings: with and without using masks. 
In the without mask case, our method produces an almost identical quality while reducing the final number of Gaussians by about 10\% on average. 
This positively impacts the training time but most significantly reduces memory requirements and the storage size of the exported model. 
The underlying representation of 2DGS does not change between the with and without mask scenario, only the output mesh. 
For this reason, the training time and number of Gaussians are the same. 
Our method with masks halves the number of Gaussians over the 2DGS. 
This improves the training time more significantly and the exported model is roughly half as large. 
Although there is a minor drop in quality compared to 2DGS, it is still exceptional compared to the state of the art shown previously.

\subsubsection{Mip-NeRF360}\label{sec:exp_m360}

\begin{figure}
    \centering
    \begin{subfigure}{0.32\linewidth}
        \includegraphics[width=\linewidth]{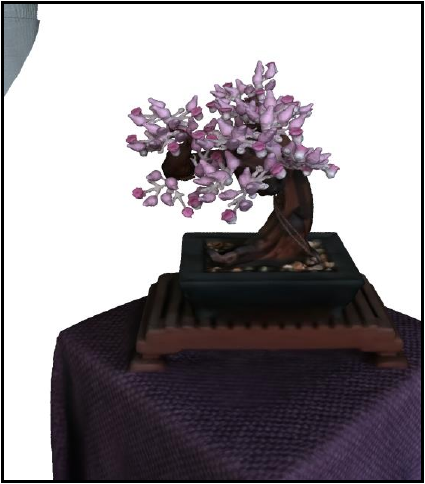}
    \end{subfigure}
    % \hfill
    \begin{subfigure}{0.32\linewidth}
        \includegraphics[width=\linewidth]{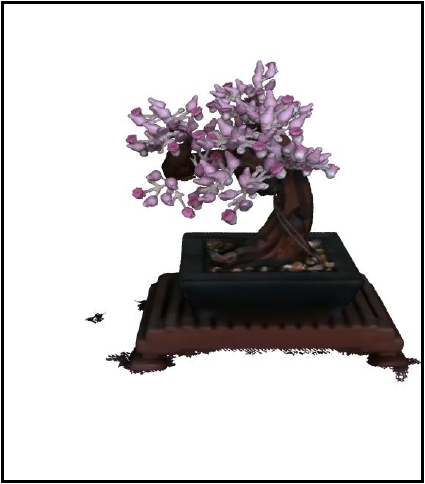}
    \end{subfigure}
    % \hfill
    \begin{subfigure}{0.32\linewidth}
        \includegraphics[width=\linewidth]{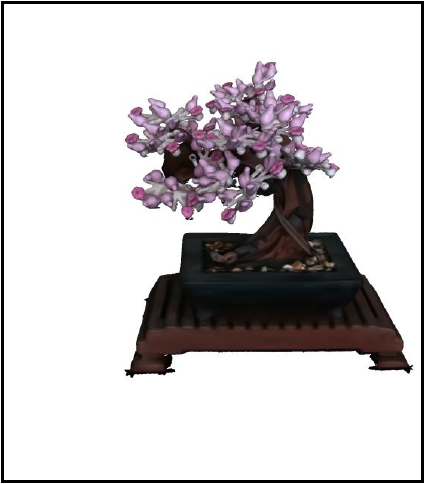}
    \end{subfigure}
    % \hfill
    \begin{subfigure}{0.32\linewidth}
        \includegraphics[width=\linewidth]{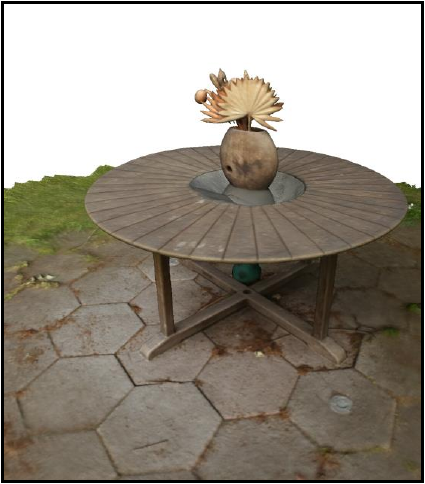}
    \end{subfigure}
    % \hfill
    \begin{subfigure}{0.32\linewidth}
        \includegraphics[width=\linewidth]{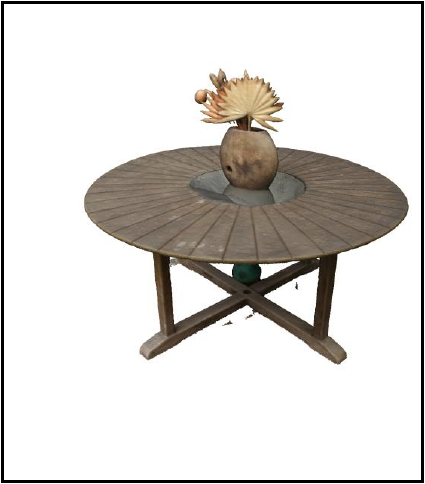}
    \end{subfigure}
    % \hfill
    \begin{subfigure}{0.32\linewidth}
        \includegraphics[width=\linewidth]{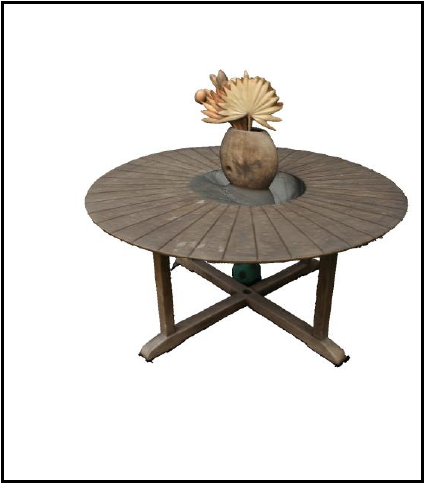}
    \end{subfigure}
    % \hfill
    \begin{subfigure}{0.32\linewidth}
        \includegraphics[width=\linewidth]{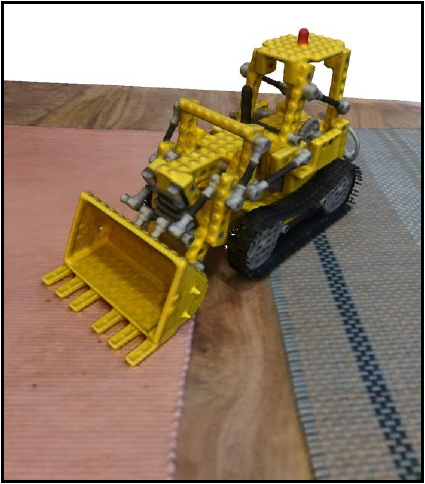}
    \end{subfigure}
    % \hfill
    \begin{subfigure}{0.32\linewidth}
        \includegraphics[width=\linewidth]{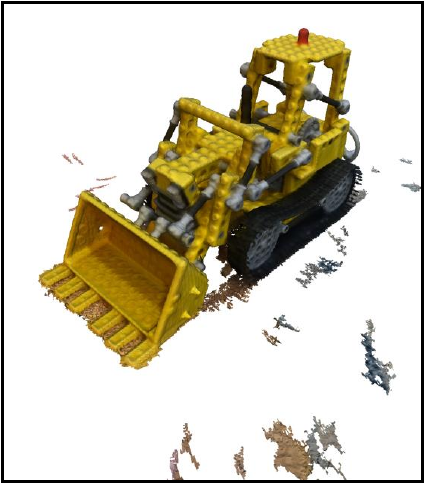}
    \end{subfigure}
    % \hfill
    \begin{subfigure}{0.32\linewidth}
        \includegraphics[width=\linewidth]{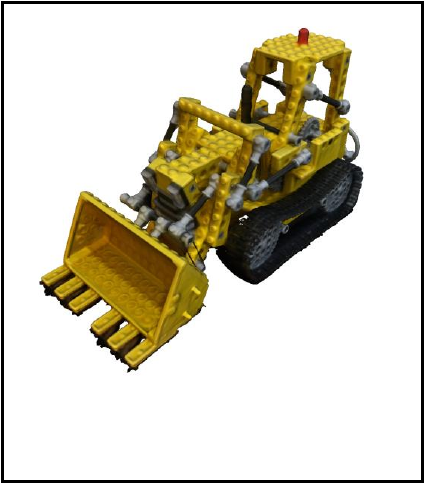}
    \end{subfigure}
    % \hfill
    % \begin{subfigure}{0.32\linewidth}
    %     \includegraphics[width=\linewidth]{figures/meshes_overview/treehill_baseline_without-mask_crop.pdf}
    % \caption{2DGS}
    % \end{subfigure}
    \subcaptionbox{2DGS}{\includegraphics[width=0.32\linewidth]{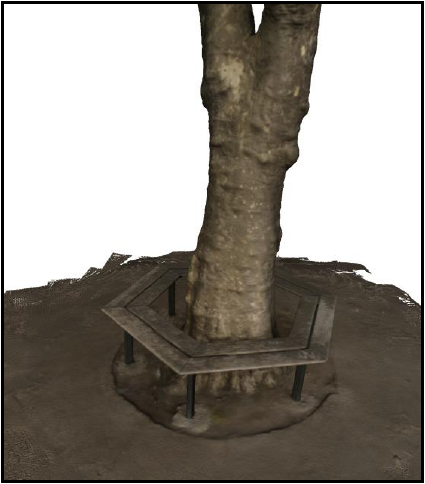}}
    % \hfill
    % \begin{subfigure}{0.32\linewidth}
    %     \includegraphics[width=\linewidth]{figures/meshes_overview/treehill_baseline_with-mask_crop.pdf}
    % \caption{2DGS + Our Masks}
    % \end{subfigure}
    \subcaptionbox{\centering2DGS +\newline Our Masks}{\includegraphics[width=0.32\linewidth]{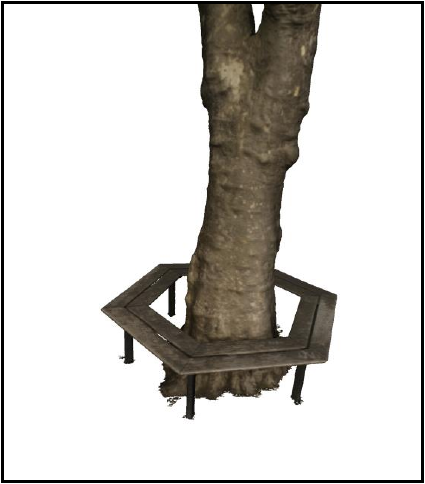}}
    % \hfill
    % \begin{subfigure}{0.32\linewidth}
    %     \includegraphics[width=\linewidth]{figures/meshes_overview/treehill_ours_crop.pdf}
    % \caption{Ours}
    % \end{subfigure}
    \subcaptionbox{Ours}{\includegraphics[width=0.32\linewidth]{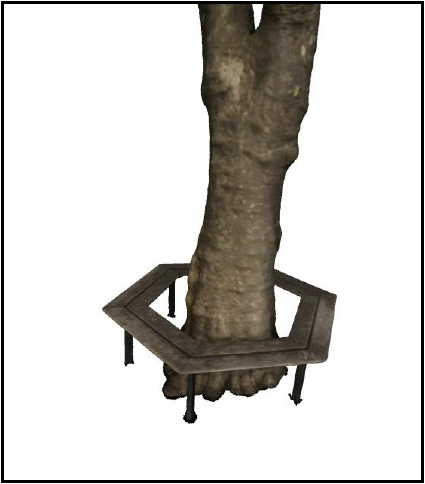}}
    \caption{Qualitative comparison of rendered meshes on the Mip-NeRF360 dataset. 2DGS meshes are extracted as bounded mesh with automatically estimated parameters. Ours is the full model extracted as unbounded mesh.}
    \label{fig:m360_meshes}
\end{figure}

The Mip-NeRF360 dataset does not have a ground truth geometry, which makes quantitative evaluation only possible for novel view rendering. 
A qualitative comparison of the meshes is shown in Figure~\ref{fig:m360_meshes}. 
We evaluate the quality of novel view rendering using the peak signal-to-noise ratio (PSNR) and the structural similarity index measure (SSIM) between the ground truth and rendered images. 
Because our method reconstructs only the necessary parts of the scene, we consider only a masked PSNR and SSIM in all our evaluations. 
The masked PSNR is given by the following equation:
\begin{equation}
    PSNR = 20 \log_{10}(\frac{MAX_I}{\sqrt{MSE}}),
\end{equation}
where $MAX_I$ represents the maximum possible intensity and $MSE$ represents the masked MSE as follows.
\begin{equation}
    MSE = \frac{1}{\sum(M)}\sum[(GT-R_c) \cdot M]^2
\end{equation}
Masking the SSIM is not straightforward because it uses windows. 
We choose to apply the mask to the ground truth and rendered image before computing the SSIM. 
In the last step, we compute the mean only over the valid pixels. 
We want to note that this will cause invalid pixels to be within the windows of some valid pixels, which will influence the score. 
The influence would be slightly positive due to the pixels outside of the mask being identical due to masking the inputs.

\begin{table}
    \centering
    \scriptsize
    \caption{Performance comparison on the Mip-NeRF360 dataset. Methods marked with $\star$ are without using masks.}
    \label{tab:M360_performance}
    \begin{tabular}{|l|c|c|c|c|}
        \hline
        & PSNR$\uparrow$ & SSIM$\uparrow$ & Time$\downarrow$ & Gaussians$\downarrow$ \\
        \hline
        2DGS$^\star$ & \sbest 28.31 & \sbest 0.875 & \tbest 31.44 min & \tbest 2,053,425 \\
        \hline
        Ours$^\star$ & \best 28.35 & \best 0.876 & \sbest 31.28 min & \sbest 1,875,546 \\
        \hline
        Ours & \tbest 25.69 & \tbest 0.822 & \best 9.15 min & \best 73,179 \\
        \hline
    \end{tabular}
\end{table}

\begin{figure}
    \centering
    {\epsfig{file = 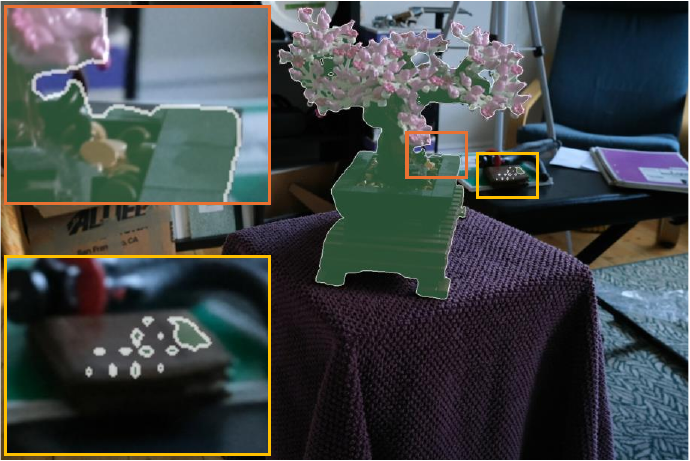, width = \linewidth}}
    \caption{Example of erroneous mask overlaid on the input image. Yellow box: non-object pixels are in mask. Orange box: some object pixels are outside mask.}
    \label{fig:erroneous_mask_overlay}
\end{figure}

\begin{figure}
    \centering
    {\epsfig{file = 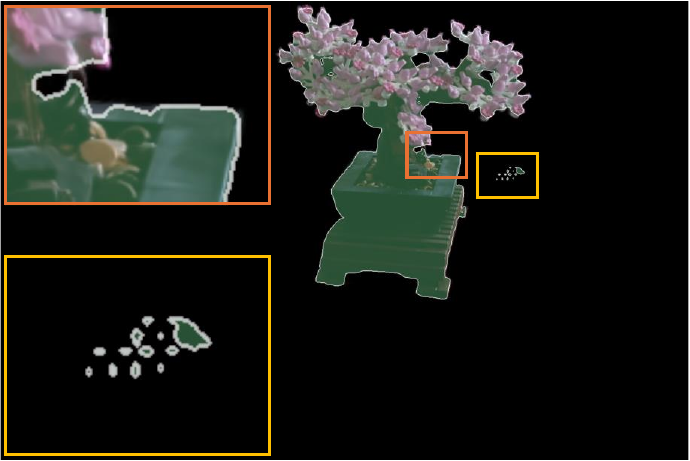, width = \linewidth}}
    \caption{Example rendering overlaid with the corresponding mask. Yellow box: non-object pixels in mask are not reconstructed. Orange box: object pixels missing in mask are still reconstructed.}
    \label{fig:robust_reconstruction_overlay}
\end{figure}

The quantitative evaluation on the Mip-NeRF360 dataset can be found in Table~\ref{tab:M360_performance}. 
Please note that the difference between 2DGS with and without masks occurs only during the mesh extraction. 
Because we evaluate the views rendered from the Gaussian representation directly, they are equivalent. 
We notice that there is no significant impact on quality from our method without masks. 
However, we are able to remove almost 10\% of Gaussians 'for free', which positively impacts training time and the storage size of the exported model. 
In the case of with masks, our method reduces the number of Gaussians by over 95\% and reduces training times by 70\%. 
Compared to the DTU dataset, the Mip-NeRF scenes are much larger and more complex. 
The size and complexity of a scene directly influence the number of Gaussians that are necessary to accurately model it. 
By using masks to focus on a specific object, we are able to discard a lot of unnecessary scene modeling. 
Although there is a small decrease in quality, we note that some errors in individual masks can cause the comparison to be biased against our method. 
Figure~\ref{fig:erroneous_mask_overlay} shows an example of an erroneous mask that includes non-object pixels. 
This negatively influences the evaluation of our method that only reconstructs the target object, while not affecting the baseline that reconstructs the entire scene. 
However, this also shows that our method is robust to minor errors in individual masks. 
Figure~\ref{fig:robust_reconstruction_overlay} shows an example rendering from our method of the same view. 
The erroneously masked part of the image is not part of the reconstruction, while the parts of the object that are missing in the mask are correctly reconstructed.

\begin{figure}
    \centering
    {\epsfig{file = 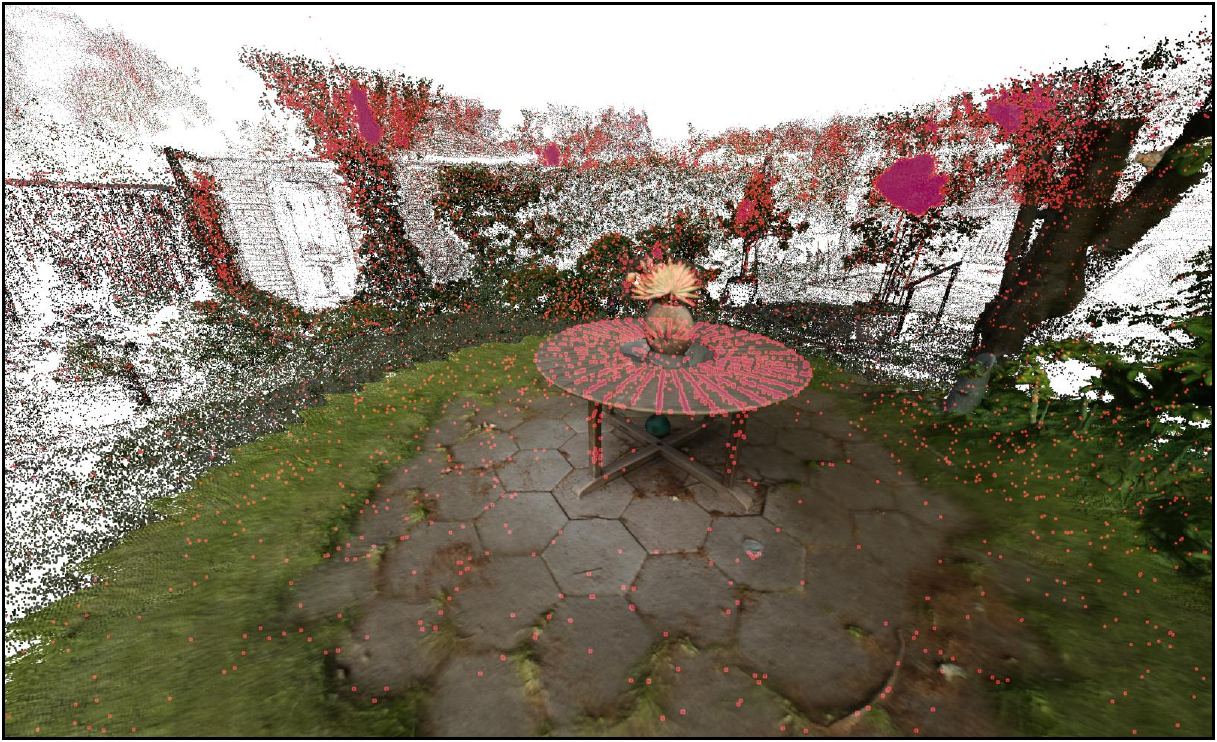, width = \linewidth}}
    \caption{Visualization of occluded Gaussians for 2DGS overlaid on the Garden scene from the Mip-NeRF360 dataset. Red areas highlight locations of occluded Gaussians.}
    \label{fig:occluded_overlay}
\end{figure}

We additionally evaluate the magnitude of occluded Gaussians by loading trained Gaussian representations and rendering all training views while tracking the Gaussian visibility. 
A visualization of occluded Gaussians for 2DGS is shown in Figure~\ref{fig:occluded_overlay}. 
The results for the Mip-NeRF360 dataset are shown in Table~\ref{tab:M360_occluded}. 
We observe that on average as many as 10\% of Gaussians in outdoor scenes are occluded. 
For indoor scenes, there are on average still close to 3\% of Gaussians occluded. 
Using our pruning approach, we are able to significantly reduce the number of occluded Gaussians to below 1\%. 
In the case of our full method including object masking, the contribution of occluded Gaussians is further reduced to below 0.3\%. 
We would like to note that the reason for any occluded Gaussians remaining in our proposed method are due to the adaptive density control being suspended in the second half of the training. 
At that point, it is still possible for Gaussians to be occluded from all views. 
However, if we continued pruning, any pruning of e.g. temporarily occluded Gaussians could no longer be fixed by densification. 
If it is desired to remove all occluded Gaussians from the final representation, it is also possible to prune them in an additional step after optimization is finished. 
This can be done in the same way as we computed the number of occluded Gaussians for this evaluation, followed by the removal of those Gaussians before exporting the final representation.

\begin{table}
    \centering
    \scriptsize
    \caption{Evaluation of occluded Gaussians on the Mip-NeRF360 dataset. Methods marked with $\star$ are without using masks.}
    \label{tab:M360_occluded}
    \adjustbox{max width=\columnwidth}{
        \begin{tabular}{|l|cc|cc|}
            \hline
            & \multicolumn{2}{c|}{Outdoor Scenes} & \multicolumn{2}{c|}{Indoor Scenes} \\
            %\hline
            & Occluded$\downarrow$ & Occluded/Total$\downarrow$ & Occluded$\downarrow$ & Occluded/Total$\downarrow$ \\
            \hline
            2DGS$^\star$ & \tbest 328088 & \tbest 9.81\% & \tbest 19333 & \tbest 2.65\% \\
            \hline
            Ours$^\star$ & \sbest 25888 & \sbest 1.00\% & \sbest 3456 & \sbest 0.49\% \\
            \hline
            Ours & \best 358 & \best 0.26\% & \best 31 & \best 0.09\% \\
            \hline
        \end{tabular}
    }
\end{table}

\subsection{Ablation}

We evaluate the design choices of our proposed method. 
To begin, we will show the effectiveness of each component as chosen by us. 
Afterward, we will consider the components individually and reason the choices.

\subsubsection{Full Model}

Table~\ref{tab:DTU_ablation} and Table~\ref{tab:m360_ablation} show that our pruning strategy alone (B) reduces the number of Gaussians and training time without impacting the quality in a meaningful way. 
It arguably is a 'free' reduction of model size and gives a small speed up. 
At the same time, our masking approach alone (C) reduces the training time and the number of Gaussians more significantly. 
However, this comes at a small loss in accuracy. 
Overall, our proposed method (D) is able to significantly reduce the training time and number of Gaussians with only a small reduction in quality. 
The loss in quality is at least partially due to masking errors as discussed in \cref{sec:exp_m360} and showcased in Figure~\ref{fig:robust_reconstruction_overlay}.

\begin{table}[hb]
    \centering
    \scriptsize
    \caption{Ablation studies comparing the influence of each of our proposed components on the DTU dataset.}
    \label{tab:DTU_ablation}
    \begin{tabular}{|l|c|c|c|}
        \hline
        & CD$\downarrow$ & Time$\downarrow$ & Gaussians$\downarrow$ \\
        \hline
        A. baseline & \best 0.748 & 10.94 min & 198,820 \\
        \hline
        B. w/ pruning & \sbest 0.750 & \tbest 10.76 min & \tbest 178,332 \\
        \hline
        C. w/ masking & 0.772 & \sbest 7.17 min & \sbest 136,561 \\
        \hline
        D. Full Model & \tbest 0.769 & \best 6.46 min & \best 108,568 \\
        \hline
    \end{tabular}
\end{table}

\begin{table}[hb]
    \centering
    \scriptsize
    \caption{Ablation studies comparing the influence of each of our proposed components on the Mip-NeRF360 dataset.}
    \label{tab:m360_ablation}
    \begin{tabular}{|l|c|c|c|c|}
        \hline
        & PSNR$\uparrow$ & SSIM$\uparrow$ & Time$\downarrow$ & Gaussians$\downarrow$ \\
        \hline
        A. baseline & \sbest 28.31 & \sbest 0.875 & 31.44 min & 2,053,425 \\
        \hline
        B. w/ pruning & \best 28.35 & \best 0.876 & \tbest 31.28 min & \tbest 1,875,546 \\
        \hline
        C. w/ masking & 25.66 & 0.821 & \best 9.12 min & \sbest 74,477 \\
        \hline
        D. Full Model & \tbest 25.69 & \tbest 0.822 & \sbest 9.15 min & \best 73,179 \\
        \hline
    \end{tabular}
\end{table}

\subsubsection{Pruning}

We show the motivation for our pruning strategy in \cref{sec:exp_m360}: 
On average between 2.65\% and 9.81\% of all Gaussians are not visible for the baseline method. 
Our pruning strategy is able to reduce the number of total Gaussians as has been shown throughout all experiments. 
Removed Gaussians representing occluded ones are signified by Table~\ref{tab:M360_occluded}, which shows that using our pruning strategy, the number of occluded Gaussians in the final representation is on average reduced by 80\% to 90\%.

\subsubsection{Background Loss}

\begin{figure}
    \centering
    \begin{subfigure}{0.48\linewidth}
        \includegraphics[width=\linewidth]{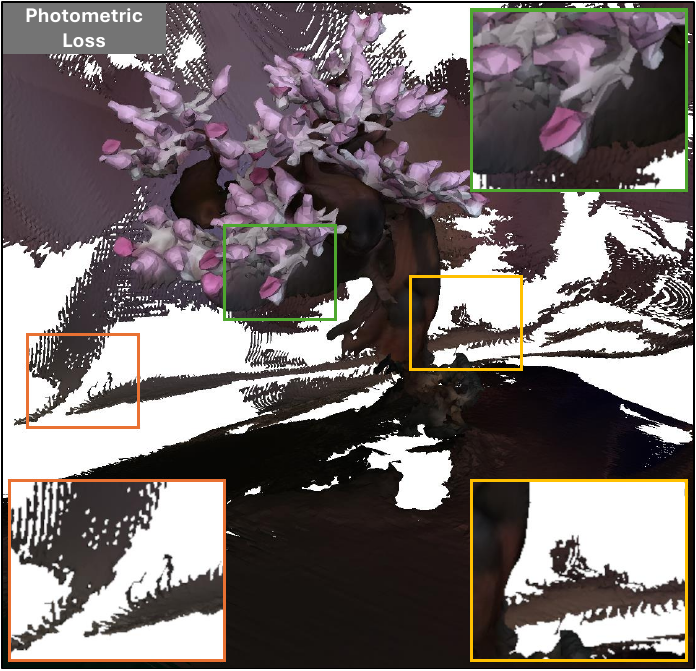}
    \end{subfigure}
    % \hfill
    \begin{subfigure}{0.48\linewidth}
        \includegraphics[width=\linewidth]{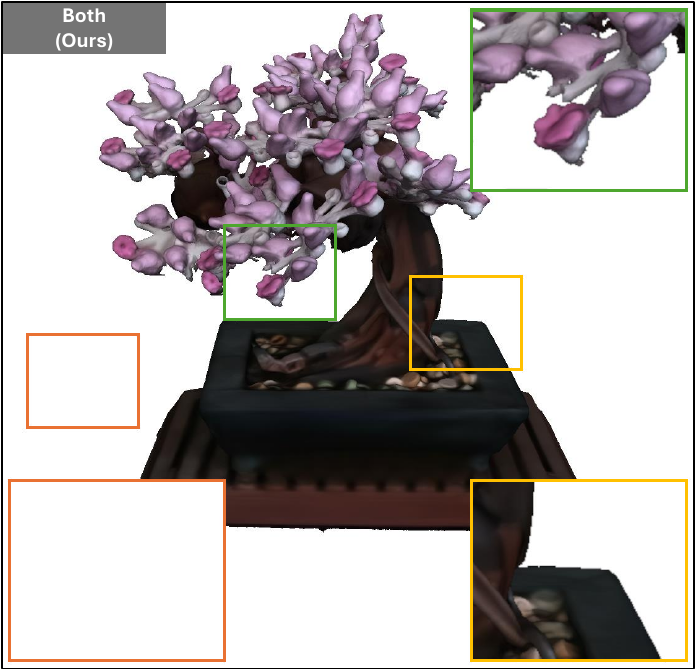}
    \end{subfigure}
    \\
    \begin{subfigure}{0.48\linewidth}
        \includegraphics[width=\linewidth]{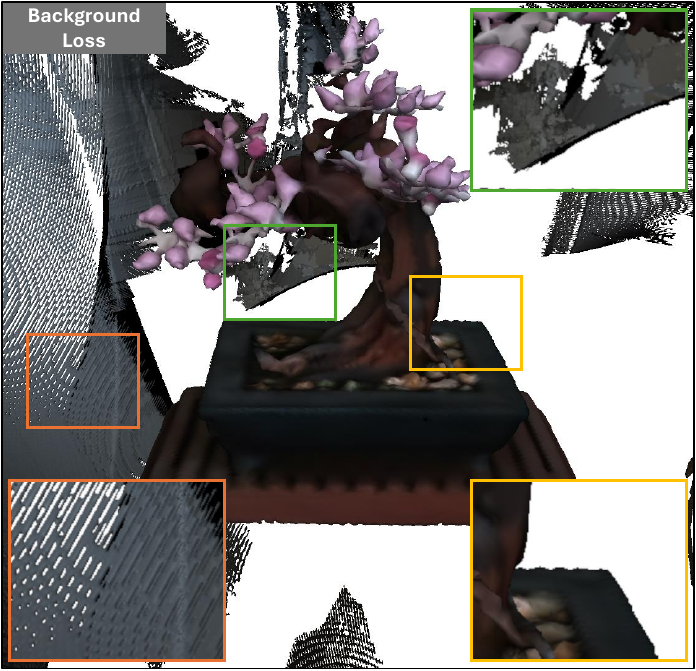}
    \end{subfigure}
    % \hfill
    \begin{subfigure}{0.48\linewidth}
        \includegraphics[width=\linewidth]{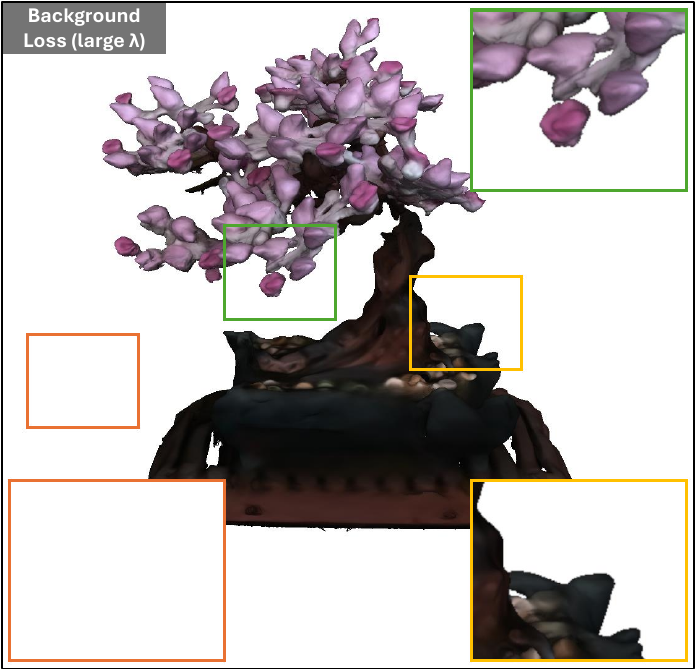}
    \end{subfigure}
    \caption{Ablation studies for background removal. Shown is masking of the photometric loss only, using our background loss only, using our background loss only with a large coefficient, and our proposed masking approach.}
    \label{fig:ablation_masking}
\end{figure}

We justify our masking approach of the background to enable object-centric reconstruction. 
Figure~\ref{fig:ablation_masking} shows a comparison of different masking approaches. 
When masking only the photometric loss, the geometry breaks and Gaussians remain in the background causing noise. 
Using only the background loss with a small lambda results in better geometry but Gaussians remain in the background, causing noise as in the previous case. 
Increasing the lambda of the background loss will eventually result in the removal of all background Gaussians, however, the geometry suffers. 
The reason for this can be twofold. 
First, is the case of occlusions, where the object is occluded by another object. 
Second, is the case of an erroneous mask, where pixels belonging to the object are incorrectly marked as not being part of the object (Figure~\ref{fig:erroneous_mask_overlay}). 
In both cases, the mask will penalize Gaussians that are correctly representing our object because the mask will indicate that the object is not visible. 
Due to the larger lambda, which is necessary to remove all background Gaussians, the penalized Gaussians would be influenced too strongly. 
Our proposed masking approach is able to properly remove background Gaussians while keeping the object's geometry intact.

% \begin{figure}
%     \centering
%     {\epsfig{file = figures/ablation_masking_crop.pdf, width = \linewidth}}
%     \caption{Ablation studies for background removal. Shown is masking of the photometric loss only, using our background loss only, using our background loss only with a large coefficient, and ours.}
%     \label{fig:ablation_masking}
% \end{figure}

\subsection{Limitations}

Our proposed method works well for the reconstruction of a range of objects but has limitations. 
First, our method inherits limitations specific to 2DGS, such as a difficulty to handle semi-transparent surfaces. 
Second, our proposed background loss depends on the quality of the input masks. 
While it is robust to some errors in individual masks, systematic errors will result in bad reconstructions. 
Additionally, the generation of masks itself is highly dependent on the characteristics of the object and the scene it is located in. 
We show an example of failure cases in \cref{appendix:failure_cases}. 
In the future, better methods for the generation of segmentation masks will alleviate this problem.

%\begin{table}[h]
%\centering
%\scriptsize
%\caption{}
%\label{tab:M360_error}
%\begin{tabular}{|c|c|c|c|c|}
%  \hline
%   & Outdoor & Scenes & Indoor & Scenes \\
%  \hline
%   & PSNR$\uparrow$ & SSIM$\uparrow$ & PSNR$\uparrow$ & SSIM$\uparrow$ \\
%  \hline
%  2DGS & \sbest 26.92 & \best 0.855 & \sbest 30.04 & \best 0.901 \\
%  \hline
%  Ours* & \best 26.97 & \best 0.855 & \best 30.07 & \best 0.901 \\
%  \hline
%  Ours & \tbest 23.95 & \sbest 0.772 & \tbest 27.87 & \sbest 0.885 \\
%  \hline
%\end{tabular}
%\end{table}

%Conclusions
\section{\uppercase{Conclusions}}

We propose an object reconstruction approach utilizing 2D Gaussians. 
Our method utilizes a novel background loss with guidance from segmentation masks. 
We are able to accurately reconstruct object surfaces even in cases of erroneous masks. 
Additionally, we propose a pruning approach that removes occluded Gaussians during training, reducing the size of the model without impacting quality. 
The object-centric reconstruction enables direct use of the learned model in downstream applications (\cref{appendix:downstream_applications}). 
Lastly, the 2D Gaussian representation is well suited for conversion to meshes. 
This enables support for applications that do not support the Gaussian representation, such as appearance editing and physics simulation for meshes.

%Acknowledgements
\section*{\uppercase{Acknowledgements}}

This work was co-funded by the European Union under Horizon Europe, grant number 101092889, project SHARESPACE. %Views and opinions expressed are however those of the author(s) only and do not necessarily reflect those of the European Union. Neither the European Union nor the granting authority can be held responsible for them. 
We thank René Schuster for constructive discussions and feedback on earlier drafts of this paper.

%References
\bibliographystyle{apalike}
{\small
\bibliography{references}}

% \renewcommand\thesection{\Alph{section}}

%Appendix
\appendix
\section*{\uppercase{Appendix}}

\section{\uppercase{Downstream Applications}}\label{appendix:downstream_applications}

\begin{figure}[h]
    \centering
    \begin{subfigure}{0.49\linewidth}
        \includegraphics[width=\linewidth]{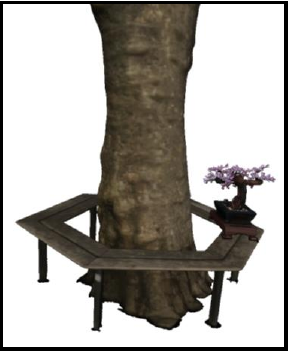}
        \caption{Mesh}
    \end{subfigure}
    \hfill
    \begin{subfigure}{0.49\linewidth}
        \includegraphics[width=\linewidth]{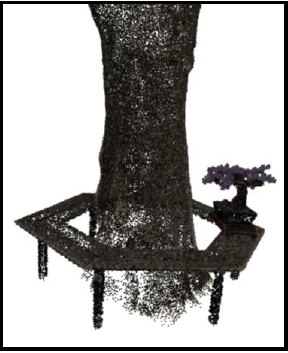}
        \caption{Gaussians}
    \end{subfigure}
    \caption{Appearance editing on the Mip-NeRF360 dataset by combining the Treehill and Bonsai scenes.}
    \label{fig:appearance_editing}
\end{figure}

Our method produces an isolated representation of a target object from the scene. 
Whether using Gaussians or mesh, the representation can be directly used without any need for additional processing. 
This enables quick and easy use for downstream applications, such as appearance editing and physics simulations. 
An example of appearance editing is shown in Figure~\ref{fig:appearance_editing}.

\section{\uppercase{Pruning in 3D Gaussian Splatting}}\label{appendix:3dgs}

We demonstrate the versatility of our pruning strategy by implementing it in 3DGS. 
We perform a single experiment on the Bicycle scene from the Mip-NeRF360 dataset as proof of principle. 
The results are shown in Table~\ref{tab:3dgs_occluded}. 
Similar to the results for 2DGS, our pruning strategy effectively reduces the number of occluded Gaussians while preserving quality. 
This reduces memory requirements and positively impacts the training time. 
However, we notice that the number of occluded Gaussians is lower compared to the 2DGS scenes. 
This is likely due to 2DGS encouraging surfaces that are fully opaque. 
The result is a clear boundary of Gaussians that are visible and Gaussians that are occluded. 
In the case of 3DGS, Gaussians can be spread out along viewing rays, which can result in semi-transparent surfaces. 
This allows for Gaussians to stay visible across different viewing angles. 
Despite the reduced magnitude of occluded Gaussians in 3DGS, we show that our pruning strategy provides a 'free' boost to the performance.

\begin{table}
    \centering
    \scriptsize
    \caption{Evaluation of our pruning strategy in 3DGS. Results on the Bicycle scene from the Mip-NeRF360 dataset.}
    \label{tab:3dgs_occluded}
    \adjustbox{max width=\columnwidth}{
        \begin{tabular}{|l|c|c|c|c|c|}
            \hline
            & PSNR$\uparrow$ & Time$\downarrow$ & Gaussians$\downarrow$ & Occluded$\downarrow$ & Occluded\%$\downarrow$ \\
            \hline
            3DGS & 25.23 & 33.1 min & 4,945,971 & 114,286 & 2.31\% \\
            \hline
            Pruning & \best 25.29 & \best 32.8 min & \best 4,830,703 & \best 9,793 & \best 0.20\% \\
            \hline
        \end{tabular}
    }
\end{table}

\begin{figure}
    \centering
    {\epsfig{file = 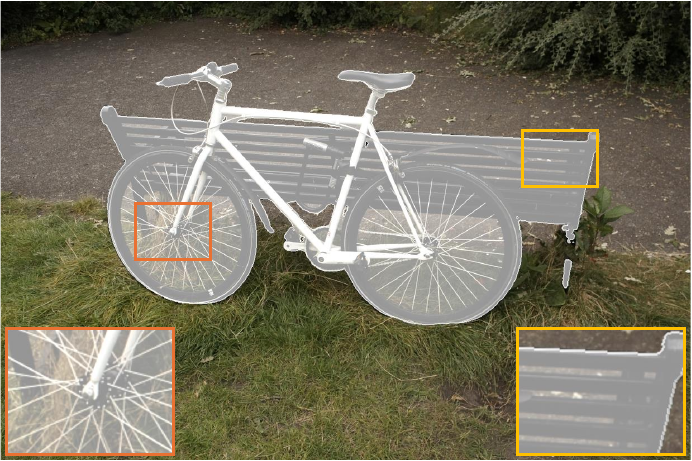, width = \linewidth}}
    \caption{Example failure case for mask generation using SAM~2. Gaps in the bench and between the bicycle spokes are wrongly included.}
    \label{fig:failure_case_mask}
\end{figure}
\section{\uppercase{Failure Cases}}\label{appendix:failure_cases}

\begin{figure}[t]
    \centering
    {\epsfig{file = 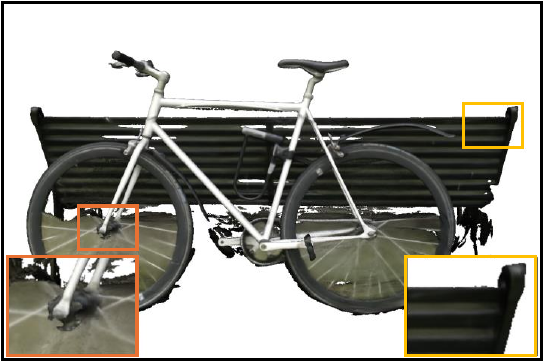, width = \linewidth}}
    \caption{Example failure case for our method with consistent errors in masks. Gaps in the bench and between the bicycle spokes are used to model the background.}
    \label{fig:failure_case_mesh}
\end{figure}

The mask generation using SAM~2 is easy and fast but the quality depends on the characteristics of the target object. 
We highlight a failure case with thin structures in the Bicycle scene of the Mip-NeRF360 dataset in Figure~\ref{fig:failure_case_mask}. 
This is a consistent error that propagates through all views of the dataset. 
The errors cause our method to learn an incorrect representation as shown in Figure~\ref{fig:failure_case_mesh}. 
Due to the masks, our method considers the gaps as part of the object, which results in a surface. 
The color represents an average of the background that is visible through the gaps. 
However, we note that our method is robust against some masking errors in individual views as shown in Figure~\ref{fig:robust_reconstruction_overlay}.

\end{document}